\documentclass[pdflatex,sn-mathphys]{sn-jnl}

\jyear{2021}%

\theoremstyle{thmstyleone}%
\theoremstyle{thmstyletwo}%

\theoremstyle{thmstylethree}%

\usepackage{makecell}
\usepackage{subcaption}
\usepackage[section]{placeins}
\usepackage{tabularx}
\usepackage{multirow}
\usepackage{CJKutf8}

\usepackage{float}
\usepackage{etoolbox}
\makeatletter
\patchcmd{\ps@headings}
{\hbox to \hsize{\hfill Springer Nature 2021 \LaTeX\ template\hfill}}
{\hbox to \hsize{}}
{}
{}
\patchcmd{\ps@titlepage}
{\hbox to \hsize{\hfill Springer Nature 2021 \LaTeX\ template\hfill}}
{\hbox to \hsize{}}
{}
{}
\makeatother
\makeatletter
\patchcmd{\ps@headings}
{\hbox to \hsize{\hfill Springer Nature 2021 \LaTeX\ template\hfill}}
{\hbox to \hsize{}}
{}
{}
\patchcmd{\ps@headings}
{\hbox to \hsize{\hfill Springer Nature 2021 \LaTeX\ template\hfill}}
{\hbox to \hsize{}}
{}
{}
\patchcmd{\ps@titlepage}
{\hbox to \hsize{\hfill Springer Nature 2021 \LaTeX\ template\hfill}}
{\hbox to \hsize{}}
{}
{}
\makeatother

\begin{document}

\begin{CJK*}{UTF8}{gbsn}

\title[FuXi-Extreme]{FuXi-Extreme: Improving extreme rainfall and wind forecasts with diffusion model}


\author[1]{\fnm{Xiaohui} \sur{Zhong}}\email{x7zhong@gmail.com}
\equalcont{These authors contributed equally to this work.}

\author[1]{\fnm{Lei} \sur{Chen}}\email{cltpys@163.com}
\equalcont{These authors contributed equally to this work.}

\author[1]{\fnm{Jun} \sur{Liu}}\email{liujun$\_$090003@163.com}

\author[1]{\fnm{Chensen} \sur{Lin}}\email{linchensen@fudan.edu.cn}

\author*[1]{\fnm{Yuan} \sur{Qi}}\email{qiyuan@fudan.edu.cn}

\author*[1]{\fnm{Hao} \sur{Li}}\email{lihao$\_$lh@fudan.edu.cn}

\affil[1]{\orgdiv{Artificial Intelligence Innovation and Incubation Institute}, \orgname{Fudan University}, \orgaddress{\city{Shanghai}, \postcode{200433}, \country{China}}}

\abstract{Significant advancements in the development of machine learning (ML) models for weather forecasting have produced remarkable results. State-of-the-art ML-based weather forecast models, such as FuXi, have demonstrated superior statistical forecast performance in comparison to the high-resolution forecasts (HRES) of the European Centre for Medium-Range Weather Forecasts (ECMWF). However, ML models face a common challenge: as forecast lead times increase, they tend to generate increasingly smooth predictions, leading to an underestimation of the intensity of extreme weather events. To address this challenge, we developed the FuXi-Extreme model, which employs a denoising diffusion probabilistic model (DDPM) to restore finer-scale details in the surface forecast data generated by the FuXi model in 5-day forecasts. An evaluation of extreme total precipitation ($\textrm{TP}$), 10-meter wind speed ($\textrm{WS10}$), and 2-meter temperature ($\textrm{T2M}$) illustrates the superior performance of FuXi-Extreme over both FuXi and HRES. Moreover, when evaluating tropical cyclone (TC) forecasts based on International Best Track Archive for Climate Stewardship (IBTrACS) dataset, both FuXi and FuXi-Extreme shows superior performance in TC track forecasts compared to HRES, but they show inferior performance in TC intensity forecasts in comparison to HRES.}


\maketitle
\section{Introduction}
Climate change is driving to an increase in the frequency and intensity of extreme weather events, including heavy rainfall and strong winds, which have significant negative impacts on human society \cite{ebi2021extreme,parmesan2022climate}. A comprehensive assessment conducted by Kotz et al. \cite{kotz2022} analyzed the impact of excessive rainfall on the gross regional product (GRP). Their study demonstrated that the rise in both the number of wet days and extreme daily rainfall leads to a substantial reduction in global macro-economic growth rates. In addition, extreme wind events disrupt wind energy production by causing wind turbines to shut down and obstructing power generation \cite{liu2023climate}. To mitigate the losses caused by such extreme events, accurate forecasting plays a critical role in providing early warnings and timely information. Numerical weather prediction (NWP) models, based on physics and atmospheric dynamics, have been widely employed for predicting various weather parameters. However, extreme events are challenging for NWP models due to their high spatial and temporal variability, as well as their association with multiple scales, ranging from small scale droplet interactions to large-scale weather systems \cite{Hess2021}. Therefore, enhancing the prediction of these extreme events should be one of the top priorities for all weather forecast agencies.

In recent years, machine learning (ML) models have gained increasing popularity as an alternative to conventional NWP models for weather forecasting. These ML models have made rapid advancements and now outperform the world's leading physics-based NWP model, specifically the high-resolution forecast (HRES) from the European Centre for Medium-Range Weather Forecasts (ECMWF) \cite{ECMWF2021,bi2022panguweather,lam2022graphcast,chen2023fuxi}. For instance, FuXi \cite{chen2023fuxi} is a cascade of ML model optimized for three consecutive forecast time periods: 0-5 days, 5-10 days, and 10-15 days, referred to as FuXi-Short, FuXi-Medium, and FuXi-Long, respectively. Developed based on 39 years of data from the ECMWF's ERA5 reanalysis \cite{hersbach2020era5}, the cascaded FuXi model can produce 6-hourly global weather forecasts at a 0.25\textdegree resolution for 15 days, achieving performance comparable to the ECMWF ensemble mean. However, ML-based weather forecasting models encounter the issue of generating unrealistically smooth predictions. This issue becomes more pronounced as the lead time increases, resulting in the underestimation of extreme weather event intensity \cite{pathak2022fourcastnet,rasp2023weatherbench}. 

Diffusion models \cite{sohldickstein2015deep,ho2020denoising,saharia2021image,Croitoru_2023} have recently attracted considerable attention in the field of computer vision due to their remarkable capability for generating highly detailed images. These models consist of two processes: a forward process, involving the gradual addition of Gaussian noise to the input data, and a reverse process, wherein the noisy data is progressively restored to the original input data. Inspired by the success of diffusion models in generating high quality images, Chen et al. \cite{swinRDM2023} developed a super-resolution model based the diffusion model. This model is capable of reconstructing high-quality forecasts with a spatial resolution of 0.25\textdegree from low-resolution forecast data at 1.40525\textdegree resolution. Furthermore, they demonstrated the diffusion model's proficiency in capturing fine-scale details. 

In this study, we introduce the FuXi-Extreme model, specifically optimized for the prediction of extreme surface variables. This is accomplished by applying a denoising diffusion probabilistic model (DDPM) \cite{sohldickstein2015deep,ho2020denoising} to recover the finer-scale details in the surface forecast data generated by the FuXi-short model in 5-day forecasts. The forecast performance is evaluated for extreme total precipitation ($\textrm{TP}$), 10-meter wind speed ($\textrm{WS10}$), and 2-meter temperature ($\textrm{T2M}$) against the ERA reanalysis dataset. Additionally, we use the International Best Track Archive for Climate Stewardship (IBTrACS) \cite{knapp2010,knapp2018} to evaluate tropical cyclone (TC) forecast performance.

\section{Dataset}

\subsection{ERA5}

FuXi-Extreme, like FuXi, is trained using the ERA5 dataset with a spatial resolution of $0.25^\circ$ and a temporal resolution of 6 hours. Model training utilizes a subset of the ERA5 dataset, spanning six years from 2012 to 2017. While the FuXi model generates forecasts for both upper-air atmospheric variables and surface variables, FuXi-Extreme is specifically trained for predicting five surface variables. These variables include $\textrm{T2M}$, the 10-meter u wind component ($\textrm{U10}$), the 10-meter v wind component ($\textrm{V10}$),  mean sea-level pressure ${MSL}$, and $\textrm{TP}$. In this paper, the ERA5 dataset in 2018 is also used in evaluating forecast performances of ECMWF HRES, FuXi, and FuXi-Extreme. 

\subsection{Tropical cyclone dataset}

We conducted assessments of TC forecasts using the IBTrACS \cite{knapp2010,knapp2018} dataset as the reference, which is provided by the National Oceanic and Atmospheric Administration (NOAA). IBTrACS combines all accessible best track datasets from around the world into a comprehensive compilation. Each track in the dataset represents a 6-hourly time series of a TC's eye location in terms of latitude and longitude coordinates, along with other relevant features at that specific time and location. In alignment with established practices for evaluating TC predictions \cite{Magnusson2021}, we evaluate all TC tracks when FuXi, FuXi-Extreme, and HRES concurrently detect a cyclone. This approach ensures that all models are evaluated using the same set of events.

To facilitate a comparison with ECMWF HRES, we used the THORPEX Interactive Grand Global Ensemble (TIGGE) \cite{bougeault2010,swinbank2016tigge} archive, which contains cyclone tracks estimated using the operational ECMWF tracker. The ECMWF TC track data, stored in XML file format, include TC tracks derived from both ECMWF HRES and ensemble forecasts. We specifically extract the HRES forecasts based the "forecast" tag.

In addition to the IBTrACS dataset, we implemented the ECMWF TC tracking method to the ERA5 dataset to extract TC tracks and intensity for TC forecast evaluations.

\section{Methodology} 

\subsection{FuXi-Extreme model} \label{fuxi}

\begin{figure}[h]
    \centering
    \includegraphics[scale=0.7]{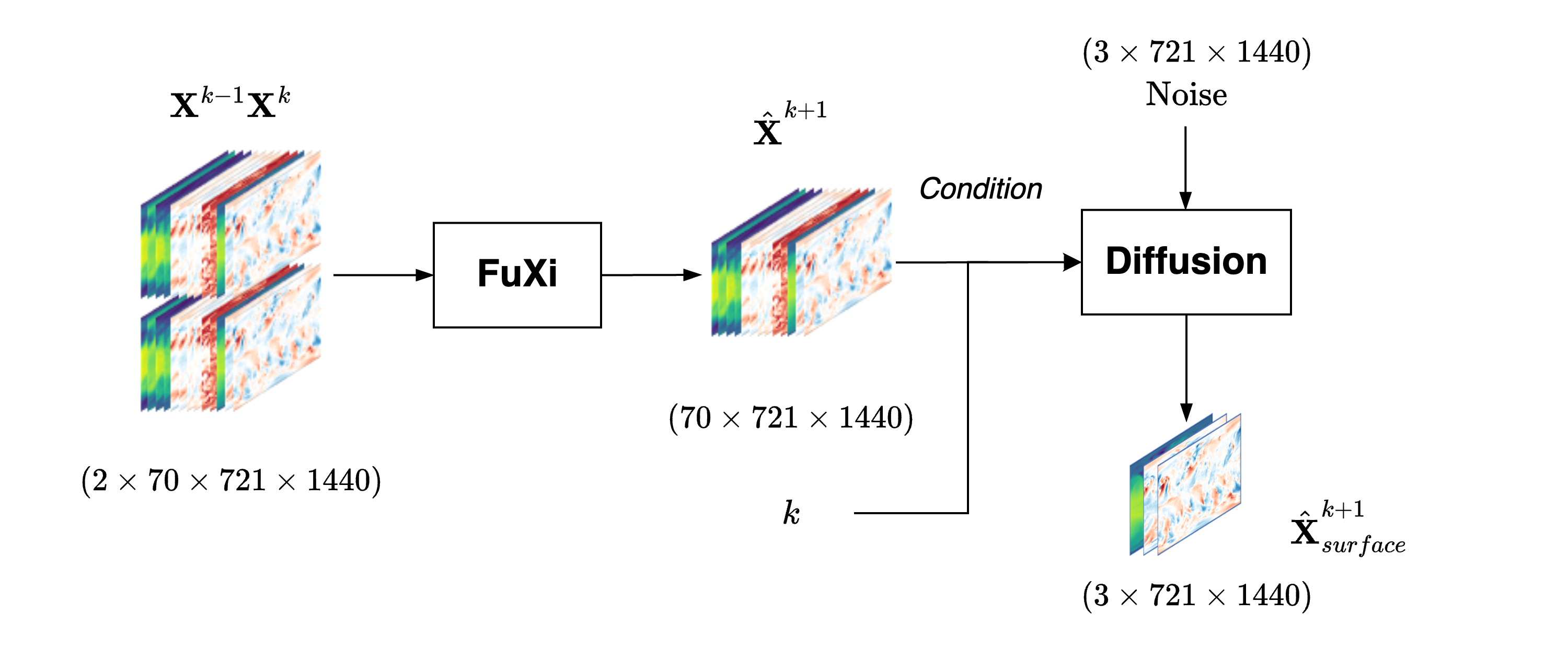}
    \caption{Schematic diagram of the structures of the FuXi-Extreme model, which consists of a FuXi model and a DDPM model.}
    \label{model}    
\end{figure}

As illustrated in Figure \ref{model}, the FuXi-Extreme is composed of a FuXi model and a DDPM model. 

\FloatBarrier

\subsubsection{Base FuXi model}
FuXi is an autoregressive model that takes input from both the previous and current time steps, denoted as $(\textbf{X}^{k-1}$ and $\textbf{X}^k)$ \footnote{$k$ denotes forecast time step.} in Figure \ref{model}, to predict the subsequent time step ($\textbf{X}^{k+1})$). The input data, including upper-air and surface variables, has dimensions $2\times70\times721\times1440$. The base FuXi model consists of three primary components: cube embedding, utilized to reduce the spatial and temporal dimensions of the input data; U-Transformer, constructed with a downsampling block, 48 repeated Swin Transformer V2 \cite{liu2022swin} blocks, and an upsampling block, designed for processing the embedded data; and a fully-connected layer that generates the final prediction. In this work, the FuXi model for 0-5 days forecasts has frozen model parameters.

\subsubsection{Denoising diffusion probabilistic model (DDPM)}

DDPM (Diffusion Probabilistic Models) is extensively utilized in image and video generation, known for its ability to create images of outstanding quality with intricate details. Notably, in tasks such as image super-resolution and image restoration \cite{saharia2021image}, DDPM excels in recovering fine-scale structures from images that may be either smooth or blurred. In this study, we utilize the DDPM model to enhance predictions produced by FuXi, with a specific focus on predicting extreme values.

DDPM models \cite{ho2020denoising} are composed of both a forward diffusion process and a reverse diffusion process. When given a data sample drawn from the real data distribution $\textbf{X}_0 \sim q(\textbf{X}_0)$, the forward diffusion process operates as a Markov chain. This process incrementally introduces Gaussian noise into the sample, following a predefined variance schedule over $T$ steps: $\beta_1, \cdots, \beta_T$. The variance schedule is critical in ensuring the effective operation of diffusion models, as it can significantly affect the quality of generated images and the convergence behavior of the model.
\begin{equation} 
\label{diffusion}
     q(\textbf{X}_t \vert \textbf{X}_{t-1}) = \mathcal{N}(\textbf{X}_t; \sqrt{1 - \beta_{t}}\textbf{X}_{t-1}, (\beta_t)\mathcal{I})
\end{equation}
where $\mathcal{I}$ represents unit variance. When the noise magnitude, $\beta_t$, added at each step is sufficiently small, and the total number of steps ($T$) is sufficiently large, the resulting output, $\textbf{X}_T$, closely approximates an isotropic Gaussian distribution. As a result, if we learn the reverse distribution $q(\textbf{X}_{t-1}\vert X_t)$, we can sample from the the Gaussian noise distribution $\mathcal{N}(0, \mathcal{I})$ (i.e., $\textbf{X}_T$), and then run the reverse process to obtain a sample from $q(\textbf{X}_0)$. However, the calculation of the posterior $q(\textbf{X}_{t-1}\vert \textbf{X}_t)$ is challenging as it requires using the entire dataset. Instead, $q(\textbf{X}_{t-1}\vert \textbf{X}_t)$ is approximated with a parameterized model, $p_\theta$. Since $q(\textbf{X}_{t-1}\vert \textbf{X}_t)$ is Gaussian, $p_\theta$ is also Guassian for sufficiently small $\beta_{t}$, and the parameterization involves the mean and variance only: 
\begin{equation} 
\label{reverse}
     p_\theta(\textbf{X}_{t-1} \vert \textbf{X}_T) =\mathcal{N}(\textbf{X}_{t-1}; \mu_{\theta}(X_t, t), \Sigma_\theta(\textbf{X}_t, t))
 \end{equation}
where $\mu_{\theta}(\textbf{X}_t, t)$ and $\Sigma_\theta(\textbf{X}_t, t)$ denote the mean and variance, respectively. 

As the combination of $q$ and $p$ closely resembles a variational autoencoder (VAE) \cite{kingma2022autoencoding}, it is possible to employ a comparable loss function that optimizes the negative log-likelihood ($log(p_\theta(\textbf{X}_0))$) of the training data. Moreover, a variational lower bound, often referred to as "Evidence lower bound (ELBO)", can be used. Ho et al. \cite{ho2020denoising} demonstrated that better results can be achieved by using a simplified objective, which involves predicting the noise at each step as follows:
\begin{equation} \label{e-prediction}
     L_t = E_{t \sim [1, T], \textbf{X}_t, \epsilon_t \sim \mathcal{N}(0, \mathcal{I})}[\|\epsilon_t - \epsilon_\theta(\textbf{X}_t, t)\|^2]
\end{equation}
where $t$ is uniformly ranges from 1 and $T$, $\textbf{X}_t \sim q(\textbf{X}_t \vert \textbf{X}_0)$ is obtained by applying Gaussian noise $\epsilon$ to $\textbf{X}_0$, and $\epsilon_\theta$ is the model to predict the added noise. This loss function is just a "mean squared error (MSE)" between the noise added in the forward process and the noise predicted by the model. 

The DDPM model incorporates all variables generated by FuXi as conditions. During training, the FuXi model predicts $\textbf{X}^{k}$ in real-time, with decreasing forecasting accuracy as forecast lead times increase. Diffusion models can be conditioned on additional inputs \cite{dhariwal2021diffusion, ho2022cascaded, whang2022deblurring, nichol2021glide, ramesh2022hierarchical}. In our case, the DDPM model is further conditioned on FuXi model's time step, denoted as $k$. This conditioning allows us to obtain a posterior:
\begin{equation} \label{posterior}
     p_\theta(\textbf{Y}^k_{t-1}\vert \textbf{Y}^k_{t}, \textbf{X}^{k}, t, k), 
\end{equation}
where $\textbf{Y}^k$ represents the ground truth corresponding to the model predicted $\textbf{X}^k$, Instead of employing the $\epsilon$-prediction approach, we predict the original targets $\textbf{X}^k$ directly, following \cite{ramesh2022hierarchical}. The model works as a denoising function, and is trained using a MSE loss:
\begin{equation} \label{x-prediction}
     L = E_{t \sim [1, T], \textbf{Y}_t^{(k)} \sim q_t}[\|\textbf{Y}^{k} - f_\theta(\textbf{Y}_t^{(k)}, \textbf{X}^{k}, t, k)\|^2]
\end{equation}

\subsection{FuXi-Extreme model training}

During the training process of the FuXi-Extreme model with Pytorch \citep{Paszke2017}, only the parameters of DDPM model are updated, while the parameters of the FuXi model remain fixed. The training process takes approximately 24 hours for 120000 iterations and is performed on a cluster equipped with 8 Nvidia A100 GPUs. A batch size of 1 is used on each GPU. Optimization is performed using the AdamW \cite{kingma2017adam,loshchilov2017decoupled} optimizer with the following parameters: ${\beta_{1}}$=0.9 and ${\beta_{2}}$=0.95, an initial learning rate of 2.5$\times$10$^{-5}$, and a weight decay coefficient of 0.1. To mitigate over-fitting, Scheduled DropPath \cite{larsson2017fractalnet} is applied with a dropping ratio of 0.2. 

\subsection{Evaluation method}

\subsubsection{Critical success index (CSI)}

CSI \cite{schaefer1990,wilks2011}, also known as the Threat Score (TS), is a widely used metric in forecast evaluations, quantifying the success of forecasts for hit rates. It is calculated using the equation \(CSI = \frac{TP}{TP + FP + FN}\) and its value ranges from 0 to 1. A value of 0 indicates no predictive skill, while a value of 1 represents the best possible score. Here, $TP$, $FP$, $FN$, and $TN$ represent true positives (hits), false positives (false alarms), false negatives (misses), and true negatives, respectively. 

To compute the CSI metrics for $\textrm{TP}$ and $\textrm{WS10}$, specific threshold values are required. For $\textrm{TP}$ accumulated over 6 hours, we reference the 24-hour precipitation values based on China’s national precipitation grading standard (GB/T 28592-2012) and divide them by 4 to obtain threshold values of 2.5, 6.25, 25, 62.5, and 70 mm for 6-hour precipitation. For wind speed categories, we adopt the Beaufort Scale, an empirical measure created by British Admiral Sir Francis Beaufort in 1805, which relates wind speed to observed conditions at sea or on land. These wind speed thresholds are set at 13.9, 17.2, 20.8, 24.5, 28.5, and 32.7 $m s^{-1}$. 

\subsubsection{Symmetric extremal dependency index (SEDI)}

Extreme events are defined as events occurring in the tails of a variable's distribution, typically in the upper tail. These events can be identified by either surpassing a specific absolute value of a physical variable or by exceeding a percentile within the climate distribution. Absolute threshold values are more relevant for assessing damage, whereas percentile-based thresholds are useful for generating scores that are more comparable across different regions and seasons. CSI metrics are computed using the absolute threshold values, irrespective of geographical locations and time. While the CSI scores are useful for assessing forecast quality, they tend to converge to zero values for extreme events \cite{stephenson2008extreme}. To address this issue, Ferro and Stephenson \cite{ferro2011extremal} developed the SEDI, which offers a measure of the association between predicted and observed extreme events. The SEDI score is calculated using the following equation:
\begin{equation} \label{SEDI}
     SEDI = \frac{logF - logH - log(1 - F) + log(1 - H)}{logF + logH + log(1 - F) + log(1 - H)}
\end{equation}
where ${F}$ is the false alarm rate (\(F = \frac{FP}{FP + TN})\)) and ${H}$ is the hit rate (\(H = \frac{TP}{TP + FN}\)). SEDI values range from -1 to 1, with 0 representing no skill and 1 indicating a perfect forecast.

Unlike the static threshold values used in calculating CSI metrics, SEDI calculations incorporate threshold values that are specific to individual locations and times. Following the methodology proposed by Magnusson et al. \cite{magnusson2014}, the reference climatology is derived separately for each variable, month of the year, time of day, and latitude/longitude coordinate. This approach improves the detection of extreme events by mitigating the impact of diurnal and seasonal cycles at individual spatial locations. To illustrate this, we can consider an example of more than 5 mm of rainfall within 6 hours, where significance varies greatly depending on geographical locations. For instance, such an event would be considered as extreme in the arid Saharan Desert but not in Big Bog, Hawaii, one of the rainiest place on Earth. In this study, the reference climatology includes the 90th, 95th, 98th, and 99.5th percentiles, derived from the ERA reanalysis data spanning the years from 1993 and 2016. As shown in Figures \ref{quantile_t2m} to \ref{quantile_ws10} in the Appendix, the percentile values exhibit significant variations across different geographical regions. Regarding $\textrm{T2M}$, the percentile values also reveal notable differences among various months of the year. 

\subsubsection{Tropical cyclone tracking method}

For both FuXi and FuXi-Extreme forecasts, we applied our re-implementation of ECMWF's TC tracker algorithm described by van der Grijn \cite{der2002tropical}.

Firstly, the tracking algorithm utilizes TC observation data to establish the initial TC position estimate. The first task of the tracker is to identify the cyclone within the initial conditions (analysis) by locating the minimum $\textrm{MSL}$ within a 445 km radius of the observed TC position. Subsequently, the tracker continues this search in the forecast data until it no longer detects a cyclone. Specifically, the tracker estimates the next TC location by updating the current estimated location with a displacement. This displacement is determined by averaging two vectors: 1) a linear extrapolation based on the past displacement between the current and prior tracked locations, and 2) an advection due to wind steering. This advection is determined by averaging the zonal ($\textrm{U}$) and meridional ($\textrm{V}$) wind components at the current TC position, considering pressure levels at 200, 500, 700 and 850 hPa, and then multiplying this average by the forecast time step ($dt$, i.e. 6 hours) ( \(\frac{\textrm{U200} + \textrm{U500} + \textrm{U700} + \textrm{U850}}{4}, \frac{\textrm{V200} + \textrm{V500} + \textrm{V700} + \textrm{V850}}{4}) \times dt\) ). 

In the initial forecast step (i.e., 6 hours forecast lead time in our case), only one estimated location is available and no linear extrapolation can be made, leading to the exclusive application of advection by wind steering. After calculating the estimate for the next TC location, the tracker assesses all local minima of $\textrm{MSL}$ within a 445 km radius of this estimate. It then seeks the candidate minima closest to the current estimate in distance that meet all three conditions outlined in Table \ref{summary}. If no minima satisfy all these conditions, or if the TC is located over high terrain, specifically above 1000 m, the tracker determines that no cyclone is present in the forecast data and terminates TC tracking. 

Furthermore, considering that FuXi-Extreme is exclusively optimized for surface variables, the tracker uses a combination of FuXi predicted upper-air atmospheric variables and FuXi-Extreme predicted surface variables in detecting TCs within the FuXi-Extreme forecast data.

\begin{table}[h]
\centering
\caption{\label{summary} An overview of the parameters, their associated levels, and the corresponding thresholds employed in the ECMWF Tropical Cyclone (TC) tracker.}
\begin{tabularx}{\textwidth}{cXXXX}
\hline
\textbf{Parameter} & \textbf{Level} & \textbf{Radius}  & \textbf{Threshold} & \textbf{Comments} \\
\hline
$\textrm{WS10}$ & Surface & 278 km & $\gt$ 8 $m s^{-1}$ & Only required over land. \\
Vorticity & 850 hPa & 278 km & absolute value $\geq 5 \times 10^{-5} s^{-1}$ \\
\makecell{Geopotential \\ thickness} & 850 hPa and 200 hPa & 278 km & & A maximum in thickness is necessary only after the TC has transitioned into an extratropical cyclone. \\
\hline
\end{tabularx}
\end{table}
\FloatBarrier

\subsubsection{TC track and intensity evaluation}
TC forecasts are evaluated in terms of both their track and intensity. Track errors are quantified using the mean absolute error (MAE), which measures distance between observed and predicted TC center positions. The evaluation of TC intensity involves assessing the maximum $\textrm{WS10}$ in the vicinity of the TC center and the $\textrm{MSL}$ at the TC centre. The forecast skills of TC intensity are assessed using the root mean square error (RMSE). The evaluation incorporates 5 TCs that occurred in 2018, provided that they are concurrently identified in the IBTrACS (or ERA5), HRES, FuXi, and FuXi-Extreme datasets.



\section{Results}

\subsection{Overall statistical performance of extreme weather forecasts}

\begin{figure}
    \centering
    \includegraphics[scale=0.3]{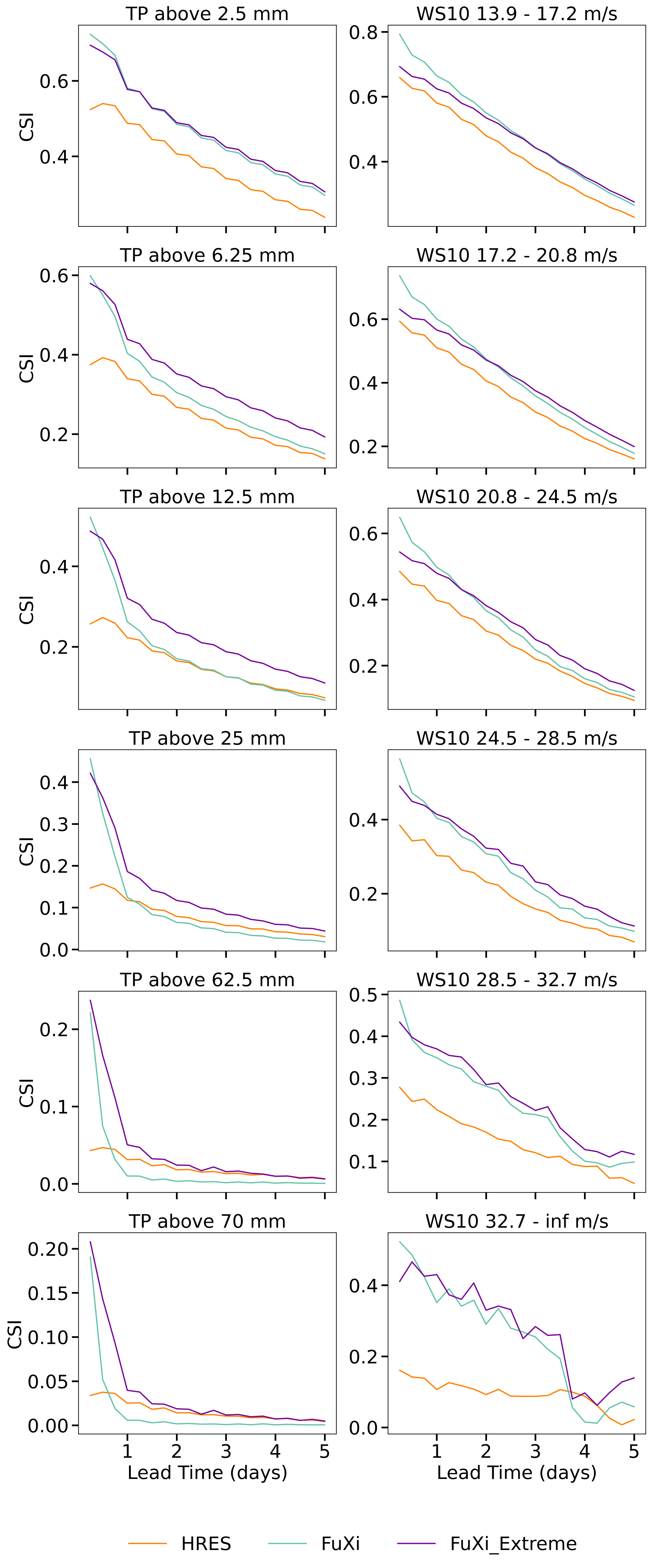}
    \caption{Comparison of CSI of the HRES (light red lines), FuXi (light blue lines), and FuXi-Extreme (light purple lines) of $\textrm{TP}$ (left column) and $\textrm{WS10}$ (right column) for various threshold values using testing data from 2018. All the forecast data are evaluated against the ERA5 reanalysis dataset.}
    \label{csi}    
\end{figure}

Figure \ref{csi} shows the time series of the CSI of HRES, FuXi, and FuXi-Extreme for $\textrm{WS10}$ and 6-hour accumulated $\textrm{TP}$ evaluated using various threshold values for 5-day forecasts. CSI values decrease as the threshold values increase for all models. The figure illustrate that, at lower threshold values, like 2.5 mm for $\textrm{TP}$, the difference in CSI between FuXi and FuXi-Extreme remains relatively small, while this difference amplifies significantly for more extreme values. Additionally, the CSI values of FuXi and FuXi-Extreme closely align at the beginning of the forecast, with FuXi-Extreme surpassing FuXi as the forecast lead time increases. Overall, FuXi-Extreme achieves the highest CSI scores for both $\textrm{WS10}$ and $\textrm{TP}$. Regarding $\textrm{TP}$, FuXi ouperforms HRES for small and moderate threshold values, but lags behind HRES for extreme thresholds, such as 62.5 and 70 mm for $\textrm{TP}$. In the case of $\textrm{WS10}$, FuXi-Extreme exhibits superior overall performance compared to FuXi and HRES. 

\begin{figure}
    \centering
    \includegraphics[width=\linewidth]{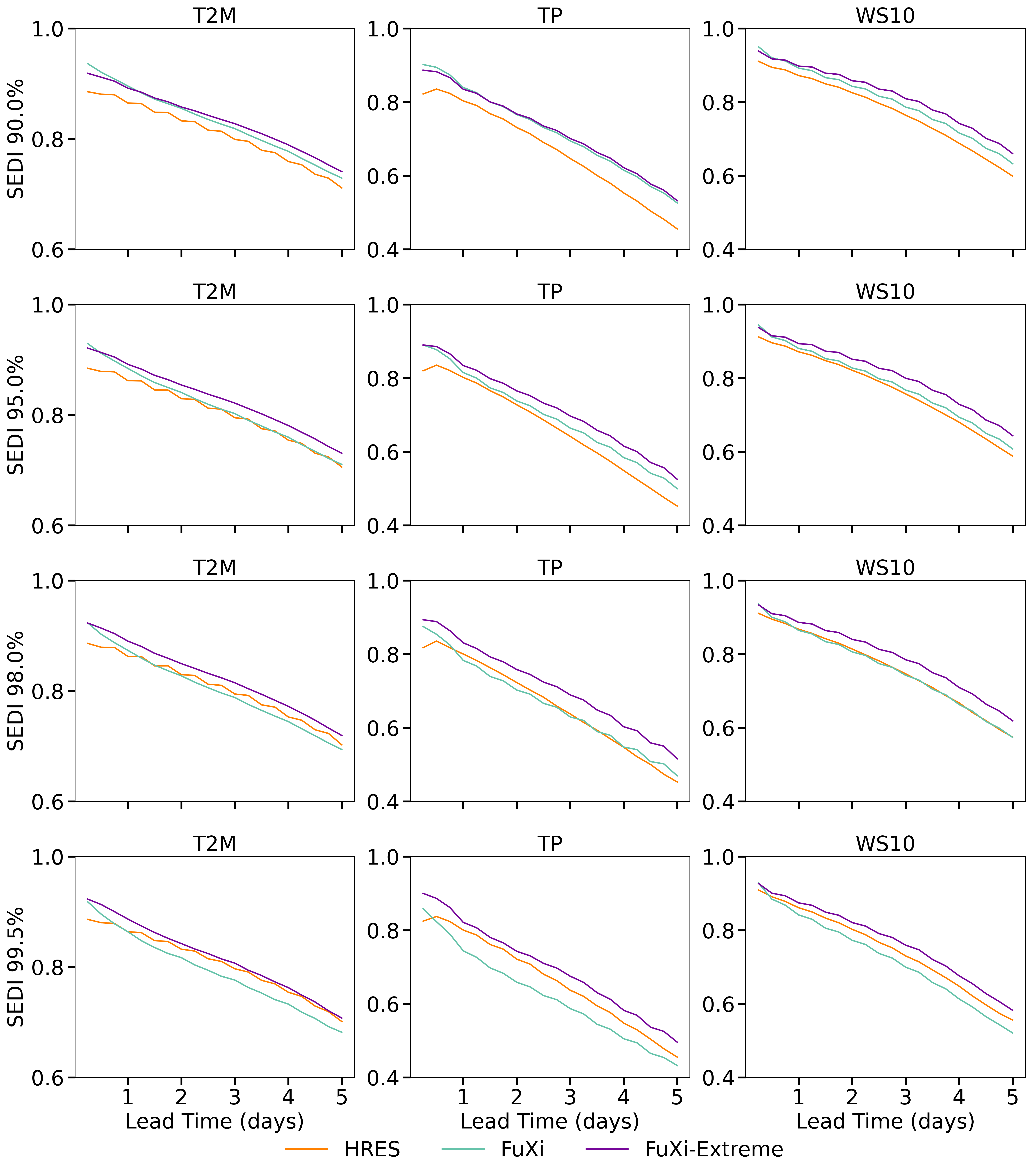}
    \caption{Comparison of SEDI of HRES (light red lines), FuXi (light blue lines), and FuXi-Extreme (light purple lines) of $\textrm{T2M}$ (left column), $\textrm{TP}$ (middle column), and $\textrm{WS10}$ (right column) for percentiles above 90th (first row), 95th (second row), 98th (third row), and 99.5th (fourth) using testing data from 2018. All the forecast data are evaluated against the ERA5 reanalysis dataset.}
    \label{sedi}    
\end{figure}

Figure \ref{sedi} presents the SEDI scores for $\textrm{T2M}$, $\textrm{TP}$, and $\textrm{WS10}$ calculated based on various percentiles for HRES, FuXi, and FuXi-Extreme, as a function of the forecast lead times in 5-day forecasts. As expected, SEDI scores decrease with increasing percentiles, indicating more challenges for more extreme events. Among the three evaluated variables, $\textrm{TP}$ shows the lowest skill, while $\textrm{T2M}$ demonstrates the highest skill, with SEDI values consistently maintaining above 0.6 throughout the entire 5-day forecast period. When comparing SEDI scores between FuXi and FuXi-Extreme, difference is negligible at the 90th percentile, but it grows substantially as the percentiles increase, indicating that FuXi-Extreme significantly outperforming FuXi at higher percentiles. FuXi surpasses HRES at the 90th and 95th percentiles, while achieving similar performance to HRES at the 98th percentile. However, for percentiles at the 99.5th, FuXi performs less accurately than HRES. Overall, FuXi-Extreme achieves the highest scores for all three variables and forecast lead times, suggesting its superior forecast capabilities for extreme $\textrm{T2M}$, $\textrm{WS10}$, and $\textrm{TP}$ in 5-day forecasts.

\begin{figure}
    \centering
    \includegraphics[width=\linewidth]{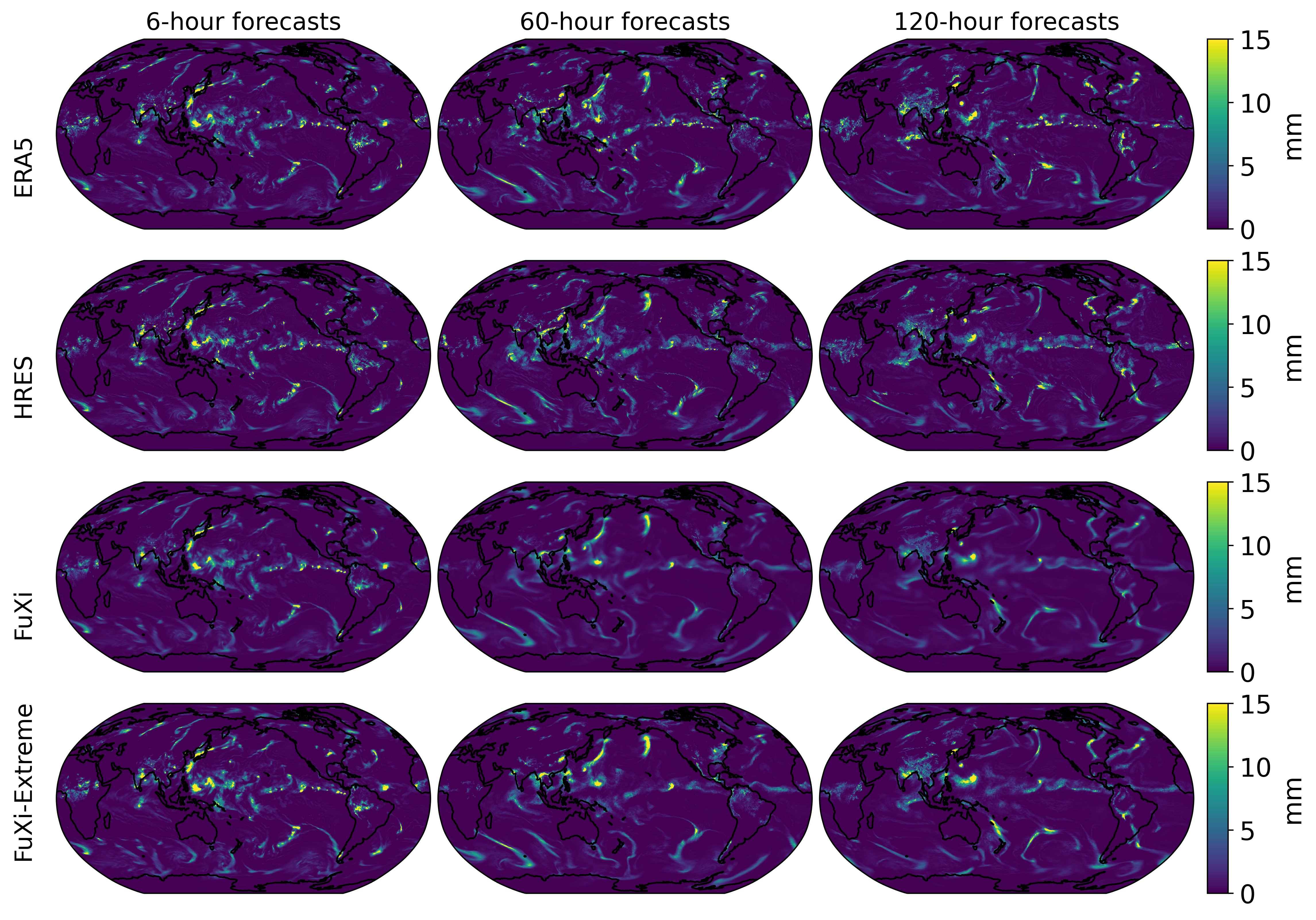}
    \caption{Comparison of snapshot examples of $\textrm{TP}$ among ERA5 (ground truth, first row), HRES (second row), FuXi (third row), and FuXi-Extrem (fourth row) for 6 (first column), 60 (second column), and 120 (third column) hours forecasts initialized at 12 UTC on August 15th, 2018.}
    \label{spatial_tp}    
\end{figure}

\begin{figure}
    \centering
    \includegraphics[width=\linewidth]{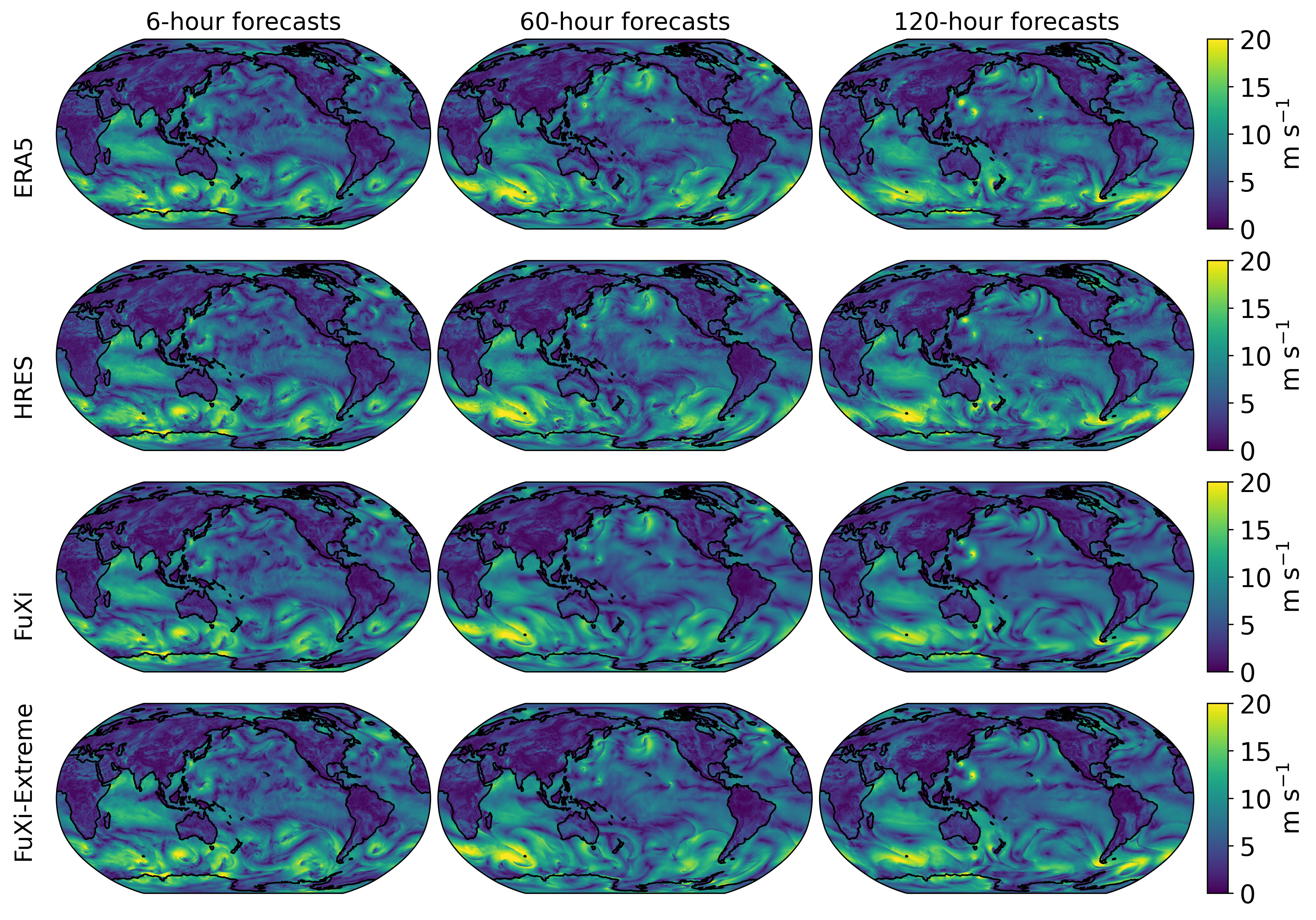}
    \caption{Comparison of snapshot examples of $\textrm{WS10}$ among ERA5 (ground truth, first row), HRES (second row), FuXi (third row), and FuXi-Extrem (fourth row) for 6 (first column), 60 (second column), and 120 (third column) hours forecasts initialized at 12 UTC on August 15th, 2018.}
    \label{spatial_ws10}    
\end{figure}

Figures \ref{spatial_tp} and \ref{spatial_ws10} show the forecast initialized at 12 UTC on August 15th, 2018. This specific time period was selected as an illustrative example due to the occurrence of Typhoon Rumbia (2018) from August 15 to 21, 2018, resulting in record-breaking extreme rainfall as it moved inland across East and Central China. The two figures reveal that FuXi-Extreme can produce forecasts with more details compared to FuXi. This distinction becomes more noticeable in the 60-hour and 120-hour forecasts as opposed to the 6-hour forecasts. Moreover, the two figures show that the HRES dataset contains finer-scale details in contrast to the ERA5 dataset, as it is derived from the original HRES dataset with a spatial resolution of 0.1\textdegree. 

\subsection{Tropical cyclone track and intensity forecast performance}

\begin{figure}
    \centering
    \includegraphics[width=\linewidth]{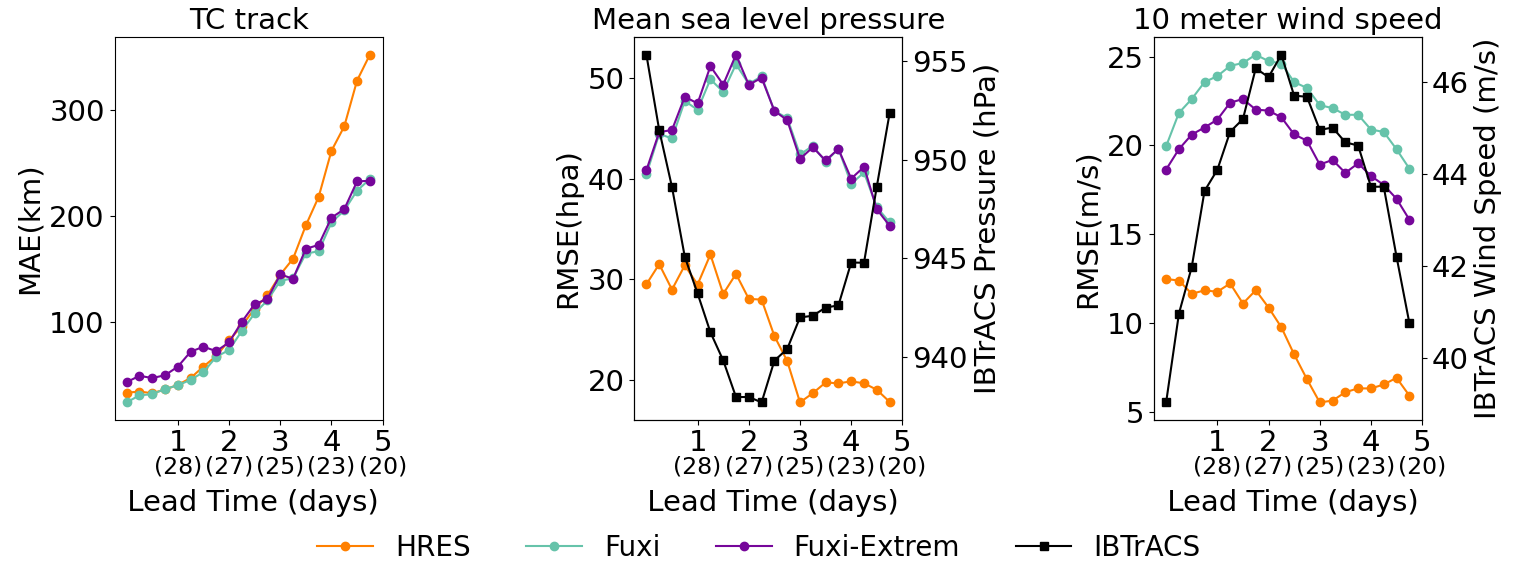}
    \caption{Comparison of the average MAE for TC track forecasts (first column) and RMSE for $\textrm{MSL}$ (second column) and $\textrm{WS10}$ (third column) in TC intensity forecasts for three models, such as HRES (light red lines), FuXi (light blue lines), and FuXi-Extreme (light purple lines), as a function of forecast lead times. The evaluation covers all TC forecasts listed in Table \ref{summary_TC}, and is performed against the IBTrACS dataset. The $\textrm{MSL}$ and $\textrm{WS10}$ forecast comparisons are dual Y axis figures, with secondary Y axis shows the IBTrACS data (black lines).}
    \label{TC_stat_IBTrACS}    
\end{figure}

Evaluation of TC forecasts includes assessments of both track and intensity predictions. In this study, we have analyzed the forecast performance for 5 TCs. Details of the specific initialization and end times of the 29 forecasts are provided in Table \ref{summary_TC}. Figure \ref{TC_stat_IBTrACS} presents a statistical comparison between HRES, FuXi, and FuXi-Extreme in 5-day forecasts, covering evaluation of both track and intensity forecasts against the IBTrACS dataset. Regarding TC track forecasts, FuXi-Extreme shows slightly inferior performance compared to HRES and FuXi for 0-2 day forecasts. Subsequently, both FuXi and FuXi-Extreme outperform HRES, and these advantages become more evident with longer forecast lead times. Notably, FuXi and FuXi-Extreme show negligible difference in track errors, as FuXi-Extreme essentially functions as an enhanced version of FuXi, designed for surface-level extreme forecasts. They share common values for upper-air variables, including U and V components and geopotential at pressure levels of 200, 500, 700 and 850 hPa, which are crucial in determining TC centers. In terms of TC intensity forecasts, FuXi-Extreme demonstrates a more noticeable improvement over FuXi in terms of the maximum $\textrm{WS10}$, as evidenced by smaller RMSE values for $\textrm{WS10}$. Regarding $\textrm{MSL}$ forecasts, the difference between FuXi-Extreme and FuXi is insignificant. HRES achieves the best performance in TC intensity forecasts, consistently maintaining the lowest RMSE values over the entire forecast period. In summary, both FuXi and FuXi-Extreme outperform HRES in TC track forecasts, while HRES surpasses FuXi and FuXi-Extreme in TC intensity predictions.

\begin{figure}
    \centering
    \includegraphics[width=\linewidth]{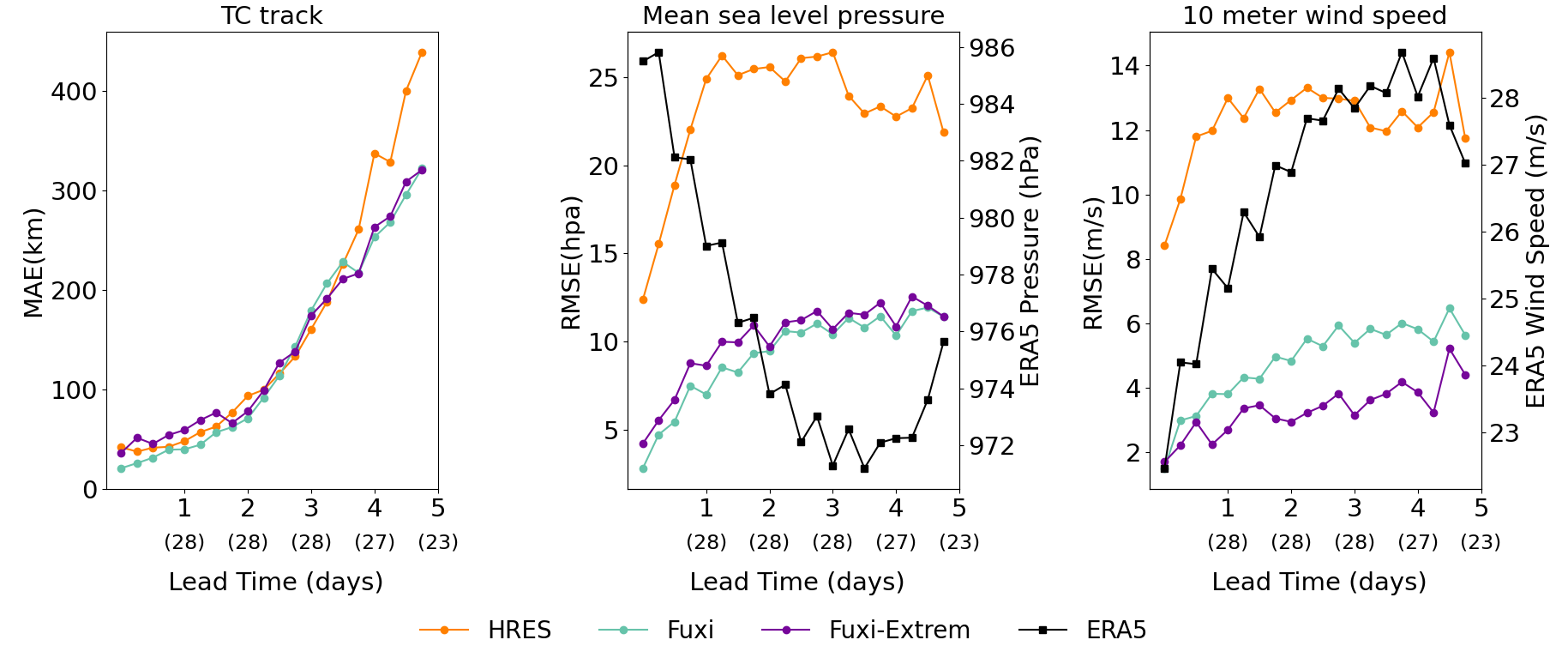}
    \caption{Comparison of the average MAE for TC track forecasts (first column) and RMSE for $\textrm{MSL}$ (second column) and $\textrm{WS10}$ (third column) in TC intensity forecasts for three models, such as HRES (light red lines), FuXi (light blue lines), and FuXi-Extreme (light purple lines), as a function of forecast lead times. The evaluation covers all TC forecasts listed in Table \ref{summary_TC}, and is performed against the ERA5 reanalysis dataset. The $\textrm{MSL}$ and $\textrm{WS10}$ forecast comparisons are dual Y axis figures, with secondary Y axis shows the ERA5 data (black lines).}
    \label{TC_stat_ERA5}    
\end{figure}

Similarly, as depicited in Figure \ref{TC_stat_IBTrACS}, Figure \ref{TC_stat_ERA5} illustrates a comparison of TC forecasts evaluated using the ERA5 dataset. The comparisons of TC track forecasts align with the previously mentioned conclusion: FuXi and FuXi-Extreme outperform HRES. In terms of TC intensity forecasts, FuXi and FuXi-Extreme show smaller RMSE values compared to HRES. This can likely be attributed to the following factors: 1) TC intensity data derived from the ERA5 dataset has smaller magnitudes than that from the IBTrACS dataset, with higher values in $\textrm{MSL}$ and lower values in $\textrm{WS10}$; 2) FuXi and FuXi-Extreme were trained using the ERA5 dataset, which is characterized by its smoothness in comparison to HRES (see Figures \ref{spatial_tp} and \ref{spatial_ws10}). The contradictory findings in comparisons of TC intensity forecasts suggest the need for a transition to a higher-resolution dataset as ground truth for training ML-based weather forecast models to enhance forecast performance \cite{Zhong_downsale_2023}.

\begin{table}
\centering
\caption{\label{summary_TC} List of TC names, their initialization times (in UTC), and end times for the 29 forecasts associated with 5 TCs evaluated in this study.}
\begin{tabular*}{\textwidth}{@{\extracolsep{\fill}} ccc}
\hline
\textbf{TC names} & \textbf{Initialized time of forecasts} & \textbf{End time of forecasts} \\
\hline
Maria & 1200 UTC 04 Jul 2018 & 1200 UTC 09 Jul 2018 \\
& 0000 UTC 05 Jul 2018 & 0000 UTC 10 Jul 2018 \\
& 1200 UTC 05 Jul 2018 & 1200 UTC 10 Jul 2018 \\
& 0000 UTC 06 Jul 2018 & 0000 UTC 11 Jul 2018 \\
& 1200 UTC 06 Jul 2018 & 1200 UTC 11 Jul 2018 \\

Rumbia & 1200 UTC 15 Aug 2018 & 1800 UTC 19 Aug 2018 \\
& 0000 UTC 16 Aug 2018 & 1200 UTC 20 Aug 2018 \\

Mangkhut & 1200 UTC 07 Sep 2018 & 1200 UTC 12 Sep 2018 \\
& 0000 UTC 08 Sep 2018 & 0000 UTC 13 Sep 2018 \\
& 1200 UTC 08 Sep 2018 & 1200 UTC 13 Sep 2018 \\
& 0000 UTC 09 Sep 2018 & 0000 UTC 14 Sep 2018 \\

KONG-REY & 1200 UTC 29 Sep 2018 & 1200 UTC 04 Oct 2018 \\
& 0000 UTC 30 Sep 2018 & 0000 UTC 05 Oct 2018 \\
& 1200 UTC 30 Sep 2018 & 1200 UTC 05 Oct 2018 \\
& 0000 UTC 01 Oct 2018 & 0000 UTC 06 Oct 2018 \\
& 1200 UTC 01 Oct 2018 & 1200 UTC 06 Oct 2018 \\
& 0000 UTC 02 Oct 2018 & 0000 UTC 07 Oct 2018 \\
& 1200 UTC 02 Oct 2018 & 1200 UTC 07 Oct 2018 \\
& 0000 UTC 03 Oct 2018 & 0000 UTC 08 Oct 2018 \\
& 1200 UTC 03 Oct 2018 & 1200 UTC 08 Oct 2018 \\

Yutu & 0000 UTC 22 Oct 2018 & 0000 UTC 27 Oct 2018 \\
& 1200 UTC 22 Oct 2018 & 1200 UTC 27 Oct 2018 \\
& 0000 UTC 23 Oct 2018 & 0000 UTC 28 Oct 2018 \\
& 1200 UTC 23 Oct 2018 & 1200 UTC 28 Oct 2018 \\
& 0000 UTC 24 Oct 2018 & 0000 UTC 29 Oct 2018 \\
& 1200 UTC 24 Oct 2018 & 1200 UTC 29 Oct 2018 \\
& 0000 UTC 25 Oct 2018 & 0000 UTC 30 Oct 2018 \\
& 1200 UTC 25 Oct 2018 & 1200 UTC 30 Oct 2018 \\
& 0000 UTC 26 Oct 2018 & 0000 UTC 31 Oct 2018 \\
\hline
\end{tabular*}
\end{table}

\section{Conclusion and future work} 

Despite significant achievements by ML-based weather forecasting models, they tend to produce increasingly smoother predictions as forecast lead times increase. These smooth predictions often result in an underestimation of extreme weather events. In this study, we developed FuXi-Extreme by training a DDPM model to enhance the granularity of surface forecasts generated by the FuXi model. Evaluations based on CSI and SEDI metrics demonstrate that FuXi-Extreme outperforms both FuXi and HRES in predicting extreme values of $\textrm{TP}$, $\textrm{WS10}$, $\textrm{T2M}$. Furthermore, evaluations of TC track and intensity forecasts show that FuXi and FuXi-Extreme outperform HRES in predicting TC tracks. However, their accuracy in TC intensity forecasts is lower than that of HRES when evaluated using the IBTrACS dataset, and they outperform HRES when evaluated against the ERA5 dataset.

While FuXi-Extreme shows promising results in forecasting extreme events, it currently focuses exclusively on surface variables. Our future efforts include augmenting our capabilities to predict extreme values for upper-air atmospheric variables after completing the development of FuXi V2.0. FuXi V2.0, designated as the second version of FuXi, which will incorporate a broader range of weather parameters. Another aspect that needs improvement is wind speed forecasts. Currently, wind speed is calculated from U and V wind components and is not directly optimized in the model training process. In FuXi V2.0, we will directly model and optimize wind speed. Moreover, visual comparisons reveal that the ERA5 dataset, commonly used as ground truth for training ML weather forecast models, is smoother compared to HRES. Therefore, we need to use a higher-resolution dataset with more details to enhance the performance of ML weather forecast models. 

Moreover, ML weather forecasting models, including FuXi and FuXi-Extreme, are purely data-driven and lack prior knowledge of the physical systems they predict. As a result, they are frequently characterized as "black boxes". While these ML models can generate accurate forecasts, their lack of transparency in the prediction process undermines confidence in their reliability. Therefore, it is crucial to interpret these ML models and verify whether their reasoning aligns with the physical understanding of the weather system. The emergence of explainable ML (XML) \cite{mcgovern2019,molnar2020,mamalakis2020,Toms2021} methods has opened up new opportunities for atmospheric research. In our future work, we plan to leverage XML methods, such as layer-wise relevance propagation (LPR) \cite{LPR2015}, to gain insights into which weather variables and input patterns are most influential and to understand why FuXi-Extreme's predictions show improvement compared to to those of FuXi for extreme events.


\section*{Data Availability Statement}
We downloaded a subset of the ERA5 dataset from the official website of Copernicus Climate Data (CDS) at \url{https://cds.climate.copernicus.eu/}. ECMWF HRES TC tracks were retrieved from the TIGGE archive in the form of downloadable XML files, which can be accessed via \url{https://confluence.ecmwf.int/display/TIGGE/Tools}. Additionally, we obtained the ground truth tracks of TC from the International Best Track Archive for Climate Stewardship (IBTrACS) project, which is publicly available at \url{https://www.ncei.noaa.gov/products/international-best-track-archive}. All the TCs in 2018 detected in FuXi and FuXi-Extreme data are available at \url{}.

\section*{Code Availability Statement}

The source code used for training and running FuXi models in this work is available at \url{https://doi.org/10.5281/zenodo.8100201} \cite{code2023}. 
The DDPM model is available at \url{lu2022dpmsolver, }.

\section*{Acknowledgements}
We express our gratitude to the researchers at ECMWF for providing the ERA5 reanalysis dataset and HRES to the research community. We acknowledged the efforts of NOAA National Centers for Environmental Information in making the IBTrACS dataset available. We also thank support from the Computing for the Future at Fudan (CFFF), which has provided us with a high-performance computing platform.

\section*{Competing interests}
The authors declare no competing interests.


\noindent


\bibliography{refs}


\end{CJK*}

\clearpage

\appendix

\renewcommand\thefigure{\thesection.\arabic{figure}}    

\setcounter{figure}{0}    

\section*{Appendix}

\section{Climatology of extreme weather events}
\label{Quantile}

\begin{figure}
    \centering
    \includegraphics[width=\linewidth]{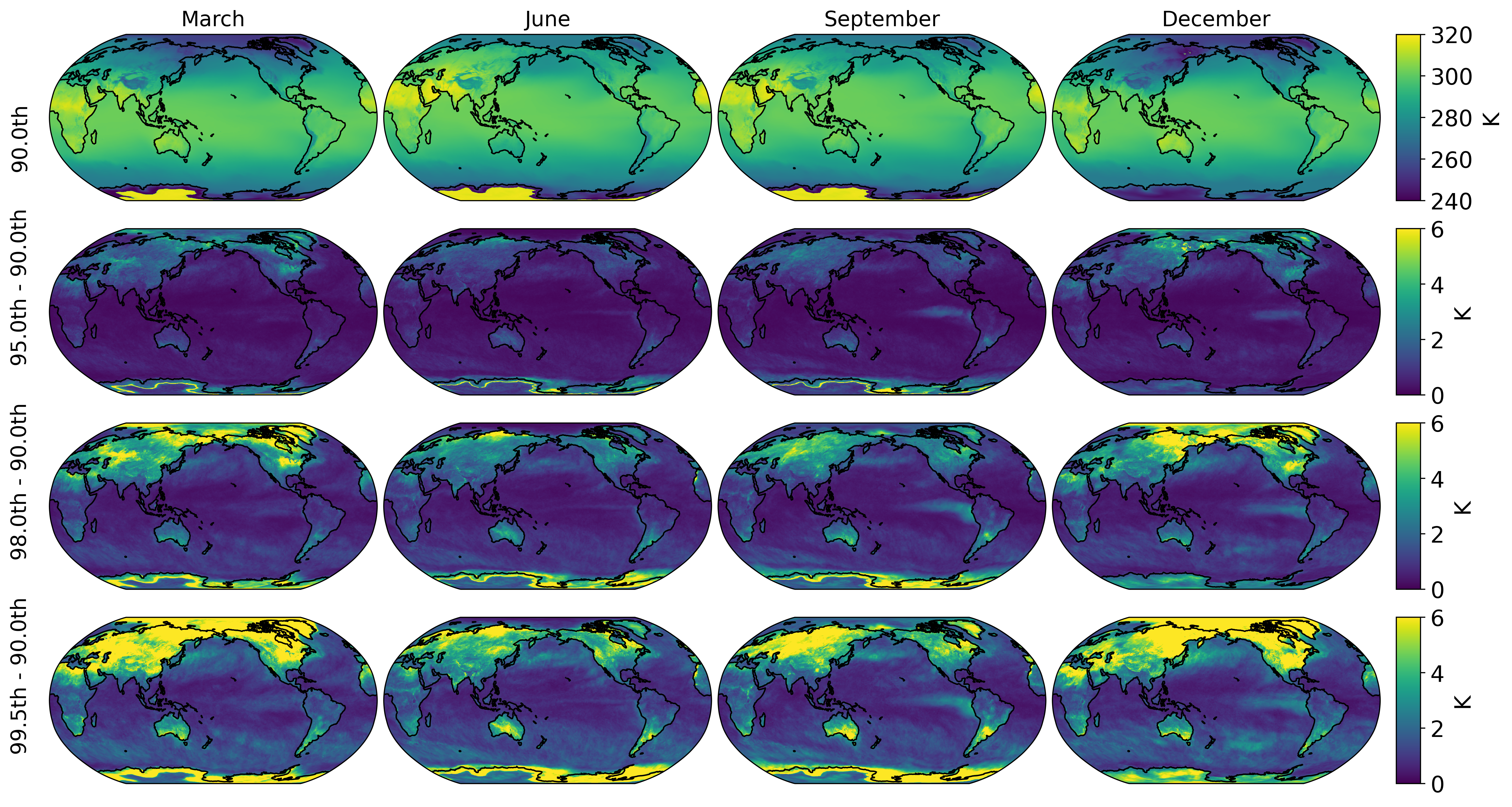}
    \caption{Values of the 90th (first row) percentile in March (first column), June (second column), September (third column), and December (fourth column), along with the difference in values between the 95th (second row), 98th (third row), and 99.5th (fourth row) percentiles and the 90th percentile for $\textrm{T2M}$.}
    \label{quantile_t2m}    
\end{figure}

\begin{figure}
    \centering
    \includegraphics[width=\linewidth]{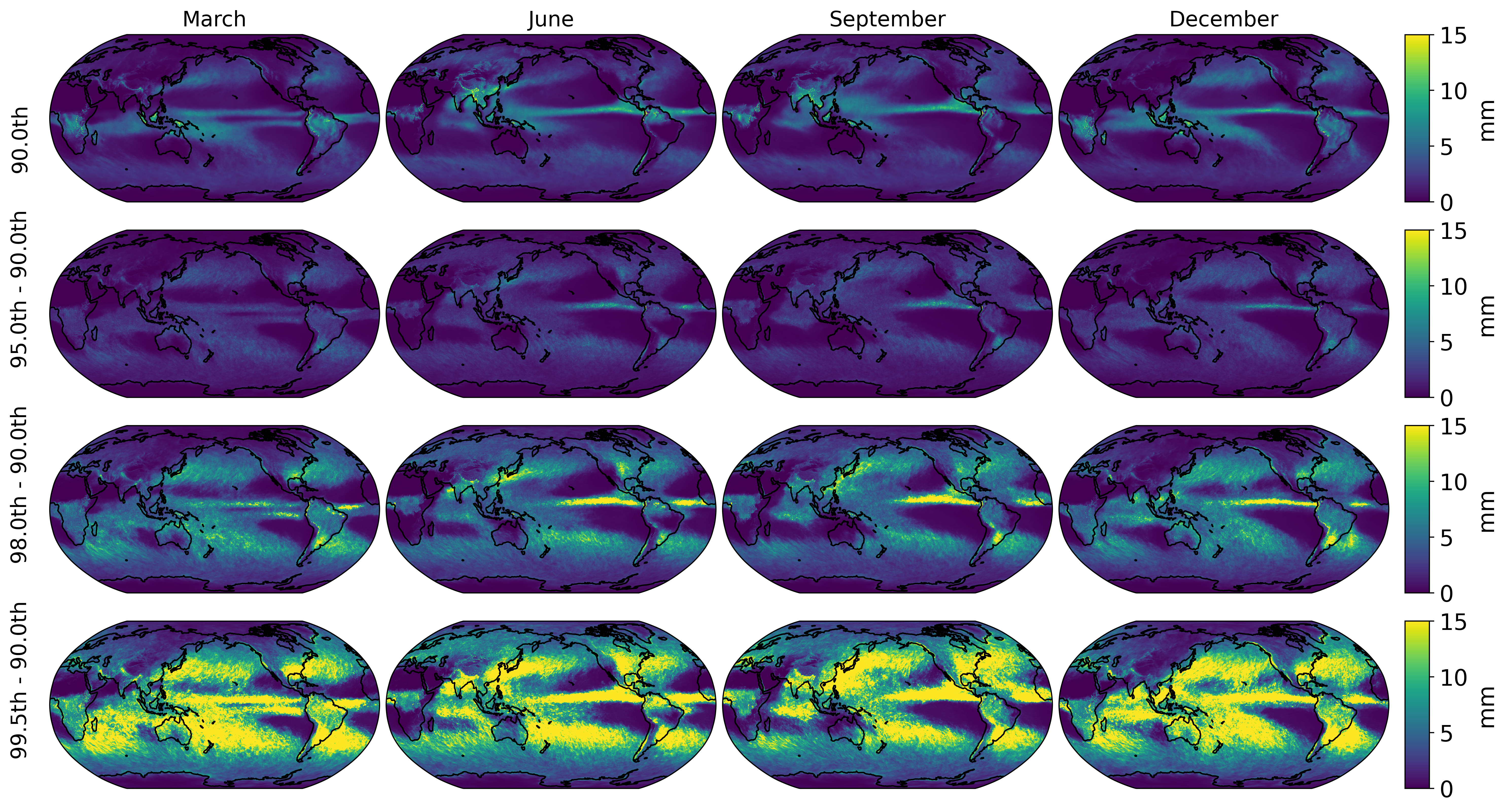}
    \caption{Values of the 90th (first row) percentile in March (first column), June (second column), September (third column), and December (fourth column), along with the difference in values between the 95th (second row), 98th (third row), and 99.5th (fourth row) percentiles and the 90th percentile for $\textrm{TP}$.}
    \label{quantile_tp}    
\end{figure}

\begin{figure}
    \centering
    \includegraphics[width=\linewidth]{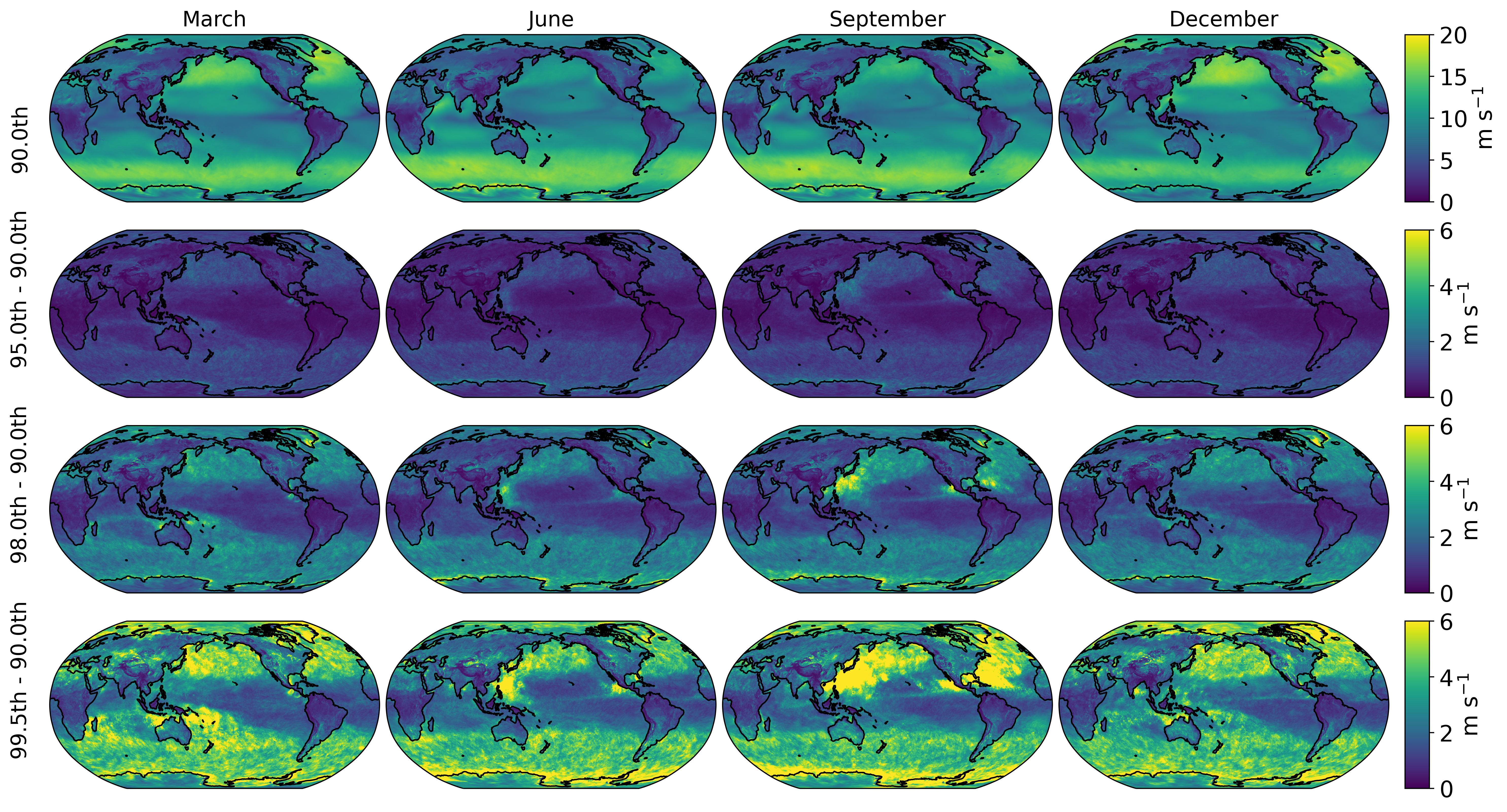}
    \caption{Values of the 90th (first row) percentile in March (first column), June (second column), September (third column), and December (fourth column), along with the difference in values between the 95th (second row), 98th (third row), and 99.5th (fourth row) percentiles and the 90th percentile for $\textrm{WS10}$.}
    \label{quantile_ws10}    
\end{figure}

Figures \ref{quantile_t2m} to \ref{quantile_ws10} show the 90th percentile from the ERA5 dataset, and the differences between the 95th, 98th, and 99.5th percentiles and the 90th percentile for $\textrm{T2M}$, $\textrm{TP}$, and $\textrm{WS10}$, respectively. The Figures illustrate significant spatial variations for all the variables. The temporal variations is also observed clearly for $\textrm{T2M}$.

\section{Overall statistical performance}
\label{rmse_acc_evaulation}

\begin{figure}
    \centering
    \includegraphics[width=\linewidth]{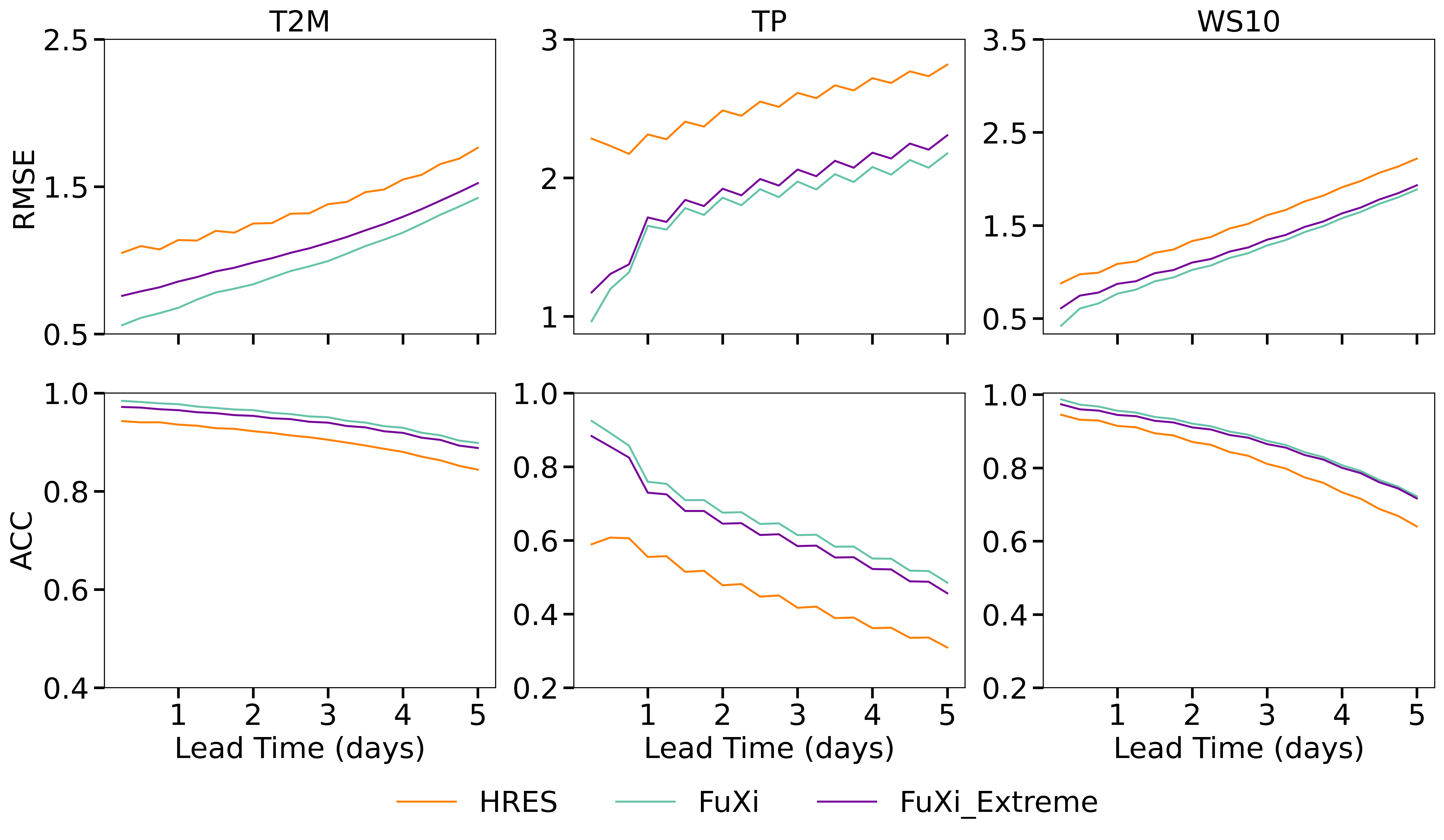}
    \caption{Comparison of the globally-averaged latitude-weighted RMSE (first row) and ACC second row) of the HRES (light red lines), FuXi (light blue lines), and FuXi-Extreme (light purple lines) of $\textrm{T2M}$ (first column), $\textrm{TP}$ (second column), and $\textrm{WS10}$ (third column) using testing data from 2018. All the forecast data are evaluated against the ERA5 reanalysis dataset.}
    \label{rmse_acc}    
\end{figure}

Figure \ref{rmse_acc} shows the time series of the globally-averaged latitude-weighted RMSE and ACC of ECMWF HRES, FuXi, and FuXi-Extreme for 3 surface variables ($\textrm{T2M}$, $\textrm{TP}$, and $\textrm{WS10}$). The figure illustrates that both FuXi and FuXi-Extreme outperform ECMWF HRES. FuXi and FuXi-Extreme have comparable performance, with FuXi slightly outperforming FuXi-Extreme, indicated by slightly lower RMSE values and the higher ACC values for all 3 surface variables.

\end{document}